\documentclass{ecai} 

\usepackage{latexsym}
\usepackage{amssymb}
\usepackage{amsmath}
\usepackage{amsthm}
\usepackage{booktabs}
\usepackage{enumitem}
\usepackage{graphicx}
\usepackage{color}
\usepackage{multirow}
\usepackage{titlesec}
\usepackage{appendix}
\usepackage{float}

\newcommand{\BibTeX}{B\kern-.05em{\sc i\kern-.025em b}\kern-.08em\TeX}
\titlespacing*{\subsection}{0pt}{3.25ex plus 1ex minus .2ex}{0.8ex plus .2ex}

\usepackage{newfloat}
\usepackage{listings}
\DeclareFloatingEnvironment[%
    listname={List of Prompts},
    name=Prompt,
    placement=htb]{prompt}

\begin{document}

\begin{frontmatter}


\paperid{7474}

\title{Probing Vision-Language Understanding through the Visual Entailment Task: promises and pitfalls}

\author[1]{\fnms{Elena}~\snm{Pitta}}
\author[1]{\fnms{Tom}~\snm{Kouwenhoven}}
\author[1]{\fnms{Tessa}~\snm{Verhoef}\thanks{Corresponding Author. Email: t.verhoef@liacs.leidenuniv.nl}} 


\address[1]{Leiden Institute of Advanced Computer Science (LIACS), Leiden University, The Netherlands}

\begin{abstract}
This study investigates the extent to which the Visual Entailment (VE) task serves as a reliable probe of vision-language understanding in multimodal language models, using the LLaMA 3.2 11B Vision model as a test case. 
Beyond reporting performance metrics, we aim to interpret what these results reveal about the underlying possibilities and limitations of the VE task.
We conduct a series of experiments across zero-shot, few-shot, and fine-tuning settings, exploring how factors such as prompt design, the number and order of in-context examples and access to visual information might affect VE performance. 
To further probe the reasoning processes of the model, we used explanation-based evaluations. 
Results indicate that three-shot inference outperforms the zero-shot baselines. 
However, additional examples introduce more noise than they provide benefits. 
Additionally, the order of the labels in the prompt is a critical factor that influences the predictions. 
In the absence of visual information, the model has a strong tendency to hallucinate and imagine content, raising questions about the model’s over-reliance on linguistic priors. 
Fine-tuning yields strong results, achieving an accuracy of $83.3\%$ on the e-SNLI-VE dataset and outperforming the state-of-the-art \mbox{OFA-X} model.
Additionally, the explanation evaluation demonstrates that the fine-tuned model provides semantically meaningful explanations similar to those of humans, with a BERTScore F1-score of $89.2\%$. 
We do, however, find comparable BERTScore results in experiments with limited vision, questioning the visual grounding of this task. 
Overall, our results highlight both the utility and limitations of VE as a diagnostic task for vision-language understanding and point to directions for refining multimodal evaluation methods. 
\end{abstract}

\end{frontmatter}

\section{Introduction}
\begin{figure}[t]
    \centering
    \includegraphics[width=0.7\linewidth]{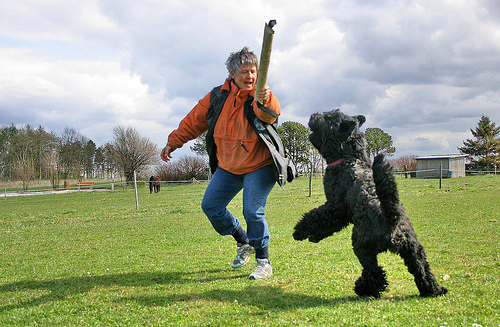}
    \caption{An example premise, hypothesis, model prediction, and explanation of the visual entailment task. \textbf{Hypothesis:} A woman carrying a stick. \textbf{Label:} \textcolor{blue}{Entailment}. \textbf{Prompt 1 prediction:} \textcolor{red}{Contradiction}. \textbf{Explanation:} "The image provides sufficient evidence to confirm that the woman is indeed carrying a stick." \textbf{Prompt 2 prediction:} \textcolor{blue}{Entailment}. \textbf{Explanation:} "The image shows a woman holding a stick, which is consistent with the description of a person carrying a stick. This suggests that the image supports or implies the truth of the hypothesis."}
    \label{fig:explanations-example2}
\end{figure}

In recent years, breakthroughs in Artificial Intelligence have driven substantial improvements in both Natural Language Processing and Computer Vision.
While these domains were traditionally separate, the emergence of multimodal learning has unified them, allowing systems to interpret, reason, and produce meaning from combined textual and visual input.
In this paper, we investigate whether a vision-language model can meaningfully combine information from visual and textual modalities in a visual entailment task. 
Visual Entailment is a multimodal task \cite{xie2019visualentailmenttaskvisuallygrounded} that extends the traditional Textual Entailment (TE) task \cite{bowman2015largeannotatedcorpuslearning, dagan2006pascal}. 
In the TE task, given a text Premise \textit{P} and a text Hypothesis \textit{H}, the goal is to determine whether a premise implies some hypothesis. 
As such, the model tested outputs a label among three possible classes: \textit{Entailment}, \textit{Contradiction}, and \textit{Neutral}, based on the relation derived from the text pair (\textit{P}, \textit{H}) \cite{dagan2006pascal, bowman2015largeannotatedcorpuslearning}. 
When there is sufficient evidence in \textit{P} to conclude that \textit{H} is true, then entailment holds. 
Wherever \textit{H} contradicts \textit{P}, a contradiction is identified. 
If not, the relation is neutral, suggesting that there is not enough data in \textit{P} to infer anything from \textit{H}. 
The difference between the TE and VE task is the replacement of the text premise with an image. 
The VE task is therefore multimodal as a model must predict by combining a visual premise with a textual hypothesis (Figure \ref{fig:explanations-example2})

This paper aims to understand the capabilities and limitations of multimodal language models, using Llama 3.2 Vision as a test case, when performing the VE task and to investigate the factors that affect its performance. 
Through a series of experiments employing zero-shot, few-shot, and fine-tuning settings, we explore the promises and pitfalls of using the VE task to probe vision-language understanding. 
Specifically, we ask: 
\begin{itemize}
    \item How does Llama 3.2 Vision perform on the visual entailment task in a zero-shot inference setting, and what is the impact of having incomplete or absent visual input?
    \item What is the impact of few-shot inference on the performance, and how does it differ with different numbers of examples?
    \item How does the order of class labels in the prompt, and the order of examples in few-shot inference affect model predictions?
    \item To what extent does fine-tuning improve model performance compared to zero-shot and few-shot inference?
\end{itemize}


Comparing the performance of Llama 3.2 Vision across these different settings, we critically reflect on how well VE truly probes vision-language understanding.

\section{Background}

Human intelligence is inherently multimodal, and learning often involves processing and integrating data from multiple senses. 
The central promise of multimodal language models is that they bring AI systems closer to human-like perception and understanding by combining the strengths of different modalities and providing a degree of language grounding \cite{baltrušaitis2017multimodalmachinelearningsurvey,liang2023foundationstrendsmultimodalmachine}.
However, there are still a number of difficulties and challenges in this domain. 
A general challenge in multimodal learning involves 
the difficulty that accompanies learning how to represent and summarize multimodal data in a way that takes advantage of the complementarity and redundancy of multiple modalities \cite{baltrušaitis2017multimodalmachinelearningsurvey}. 
Furthermore, transference \cite{liang2023foundationstrendsmultimodalmachine}, also called co-learning \cite{baltrušaitis2017multimodalmachinelearningsurvey}, referring to the ability to transfer knowledge between different modalities to aid the target modality, is still a core challenge. 
In models that combine vision and language, more specific limitations have been identified. 
For example, early work in Visual Question Answering (VQA) identified a heavy reliance on language priors where models often ignored visual input \cite{goyal2017making}. 
Similarly, it was demonstrated that VQA models may perform well by exploiting dataset shortcuts rather than truly grounding answers in the visual content \cite{jabri2016revisiting, agrawal2018don}. 
When evaluated on basic spatial relations (e.g., distinguishing "left of" vs "right of"), pre-trained models perform barely above chance, demonstrating their inability to represent spatial language robustly \cite{kamath-etal-2023-whats, shiri-etal-2024-empirical}. 
In addition, vision-language models often struggle to correctly interpret interactions among objects and their attributes, and fail to visually distinguish pairs like "a red ball on a blue cube" and "a blue ball on a red cube" \cite{Thrush_2022_CVPR, diwan-etal-2022-winoground}, showing their poor visio-compositional reasoning abilities. 
Recent investigations, moreover, find conflicting evidence regarding the presence of human-like cross-modal associations in vision-language models \cite{ alper2023bouba, verhoef-etal-2024-kiki, kouwenhoven2025crossmodalassociationsvisionlanguage}.

Even when models appear to perform well on complex multimodal tasks, this does not necessarily mean that they are reasoning in a human-like way. 
Fair evaluation should therefore not dismiss mechanistic strategies of AI models like LLMs or vision-and-language models that differ from those present in humans \citep{milliere2024anthropocentric} as they may rely on shortcut learning---the exploitation of spurious correlations or cues that happen to be present in a given dataset \cite{mitchell2023debate}.
In Computer Vision, models may latch onto dataset artifacts in images during training, making it seem like a classifier was successfully trained, for example, to distinguish between horse and non-horse images.
However, further analysis may reveal the model does not focus on horses in the images, but uses cues like copyright watermarks that only appeared in horse-labeled images \cite{lapuschkin2019unmasking}. 
In Natural Language Inference, models may perform well on benchmarks by exploiting syntactic heuristics, rather than actually understanding sentence meaning or logic \cite{mccoy-etal-2019-right}. 
Prior work has demonstrated that the Textual Entailment (TE) task, which acts as the precursor to VE, can largely be solved through the use of simple rules, such as assuming entailment if all the words in the hypothesis appear in the premise, rather than performing true semantic inference. 
Models that performed well on the standard TE dataset failed dramatically on a carefully constructed novel dataset with examples that can not be solved through sole reliance on these heuristics \cite{mccoy-etal-2019-right}.

These limitations highlight a critical disconnect between task success and genuine understanding, and demonstrate the need for evaluations that go beyond standard performance metrics. 
Here, we comprehensively compare Llama 3.2 Vision performance across multiple experiments to assess whether the VE task, as evaluated on the e-SNLI-VE \cite{kayser2021evildatasetbenchmarknatural} dataset, effectively measures multimodal understanding.

\section{Related Work}

The VE task was introduced by \citet{xie2019visualentailmenttaskvisuallygrounded}, who proposed a model called \textbf{E}xplainable \textbf{V}isual \textbf{E}ntailment model (EVE). 
This model uses attention mechanisms to learn the inner relationships in both image and text feature spaces, and achieves better performance compared to other VQA-based models.

A major advancement in this field came with the OFA model (\textbf{O}ne \textbf{F}or \textbf{A}ll) \cite{wang2022ofaunifyingarchitecturestasks}. 
OFA is a sequence-to-sequence learning framework and unifies various unimodal and cross-modal tasks, including the VE task. 
OFA achieves the state-of-the-art performance for the VE task on the SNLI-VE dataset (described in more detail in section \ref{sec:dataset}) with an accuracy of $91.2\%$ on the test set.
Extending this, OFA-X \cite{plüster2023harnessingpowermultitaskpretraining} is a proposed multitask framework that predicts not only the labels but also explanations. 
OFA-X is a fine-tuned version of the OFA model and achieved the state-of-the-art performance for the VE task on the larger e-SNLI-VE dataset (also described in more detail in section \ref{sec:dataset}) with an accuracy of $80.9\%$ on the test set. 


Perhaps the boldest perspective comes from an approach in which the proposed model CLOSE (\textbf{C}ross moda\textbf{L} transfer \textbf{O}n \textbf{S}emantic \textbf{E}mbeddings) can achieve a comparable performance, without images, using only textual input \cite{gu2023icantbelievetheres}. For the VE task, CLOSE uses the SNLI dataset for training (it uses a text premise instead of an image), while for evaluation, the SNLI-VE dataset was used, which combines vision and language. 
Despite not using images, CLOSE achieves similar performance to the image model. 
This suggests that the SNLI dataset may contain sufficient evidence to conclude the relationship without relying heavily on visual information.
This raises questions about whether a visual grounding is required and hints at the previously mentioned concept of shortcut learning \cite{mitchell2023debate}. 

The knowledge from these previous works directly influenced the design of our experiments. Inspired by OFA, we adopted a prompt-based few-shot setup to investigate how effective a model performs without direct supervision. 
In addition, the idea of explanation generation in OFA-X led us to design an experiment to analyze the explanations from the model, helping assess its interpretability and reasoning. 
Finally, the innovative approach of the CLOSE model and its findings led us to test different experiments with limited vision to explore the extent to which our model depends on visual input. 
Comparing the performance of one state-of-the-art multimodal language model, Llama 3.2 Vision, across all these different settings allows us to analyze the suitability of the VE task to probe vision-language understanding.

\section{Methodology}
In this study, we evaluate the Llama 3.2 Vision 11B model on the e-SNLI-VE dataset using three approaches: zero-shot inference, few-shot inference, and fine-tuning.

\subsection{Llama 3.2 Vision 11B}
Llama 3.2 Vision\footnote{\url{https://ollama.com/library/llama3.2-vision}} is a powerful multimodal large language model, available in two sizes: 11B and 90B parameters. The architecture of the model is based on the combination of the Llama 3.1 8B with a separately trained vision adapter \cite{huggingface2025llama32}. 
During the training phase, the text model was frozen in order to preserve text-only performance \cite{huggingface2025llama32}. 
The model was trained on 6 billion image-text pairs with a diverse data mixture \cite{huggingface2025llama32}. 
Indicatively, the 11B parameter model achieved $75.2\%$ accuracy on the VQAv2, a general visual question answering benchmark, $91.1\%$ on AI2 Diagram, a diagram understanding benchmark and $51.5\%$  on MathVista (testmini), a mathematical reasoning benchmark\footnote{\url{https://huggingface.co/meta-llama/Llama-3.2-11B-Vision}}.

\subsection{Dataset}
\label{sec:dataset}
The most common dataset used for the VE task is SNLI-VE (\textbf{S}tanford \textbf{N}atural \textbf{L}anguage \textbf{I}nference Corpus - \textbf{V}isual \textbf{E}ntailment)\footnote{\url{https://github.com/maximek3/e-ViL/tree/main/data}}. 
Specifically, this dataset is a combination of the SNLI (\textbf{S}tanford \textbf{N}atural \textbf{L}anguage \textbf{I}nference Corpus) and Flickr30k (image captioning dataset), where the premises from the SNLI are replaced with the corresponding images from Flickr30k \cite{xie2019visualentailmenttaskvisuallygrounded}. 
This was feasible because the SNLI dataset was originally built using captioned images from the Flickr30k dataset, so textual premises in SNLI could be directly matched to the caption sentences of those photos \cite{kayser2021evildatasetbenchmarknatural}. 

Although the SNLI-VE dataset is the most common dataset for the VE task, recent research documented that $39\%$ of the neutral labels in the validation and test sets were incorrectly labeled \cite{kayser2021evildatasetbenchmarknatural}. 
This happened mainly due to the replacement of the text premise with the image premise, which led to labeling errors, as an image typically contains more information than a single caption describing it \cite{kayser2021evildatasetbenchmarknatural}. 
Hence, the e-SNLI-VE (\textbf{E}xplainable \textbf{SNLI} - \textbf{V}isual \textbf{E}ntailment) dataset was created by merging SNLI-VE and e-SNLI (\textbf{E}xplainable \textbf{SNLI}). 
This yielded a visual entailment task with explanations in natural language. 
This specific dataset has better quality annotations due to hand-relabeling of validation and test sets.
The e-SNLI-VE dataset has over 430k instances. Table \ref{table:esnlive_stats} shows the dataset splits and the number of occurrences for each class in the sets. The dataset demonstrates a class imbalance, with contradiction being the most frequent class, followed by entailment with a slightly smaller number of occurrences, and neutral with the fewest cases (Table \ref{table:esnlive_stats}).
While the e-SNLI-VE dataset provides explanations, the majority of our experiments focused only on classification. 
For the experiments in which explanations were considered, this is explicitly mentioned.

\begin{table}[t]
\caption{Overview of the e-SNLI-VE dataset of \protect\citet{do2020esnlive20}.}
\centering
    \begin{tabular}{l|c|c|c}
    \toprule
    \textbf{Split}      & \textbf{Train}    & \textbf{Dev}  & \textbf{Test} \\ 
    \midrule
    \# Images           & 29,783            & 1,000         & 1,000         \\ 
    \# Entailment       & 131,023           & 5,254         & 5,218         \\ 
    \# Neutral          & 125,902           & 3,442         & 3,801         \\ 
    \# Contradiction    & 144,792           & 5,643         & 5,721         \\ 
    \# Total Labels     & 401,717           & 14,339        & 14,740        \\ 
    \bottomrule
    \end{tabular}
\label{table:esnlive_stats}
\end{table}

\subsection{Experiments}
\subsubsection{Experiment 1: Zero-shot Inference}
\label{sect:exp1}
To establish a baseline and test how well Llama 3.2 Vision can perform VE without any additional training, we first test the model in a zero-shot setting.  Here, the model is prompted to classify the image-hypothesis pair based on its pre-trained knowledge only. 
We used the prompt displayed in Prompt \ref{fig:prompt} to probe the model.

\begin{prompt}[ht!]
\scriptsize
\begin{lstlisting}[mathescape=true,escapeinside={*@}{@*}]
Perform a visual entailment classification. You are provided with two inputs:
1. Premise: An image described as follows (attached below).
2. Hypothesis: A text description.
    
Your task is to classify the relationship between the Premise (image) and Hypothesis (text) into one of the following three categories:
- Entailment: The image provides enough evidence to conclude that the Hypothesis is true.
- Contradiction: The image contradicts the Hypothesis.
- Neutral: The image does not provide enough information to determine the truth of the Hypothesis.

Provide a single classification in your response: one of Entailment, Contradiction, or Neutral. Do not include explanations, commentary, or any additional text in your response.

[Insert hypothesis]
[Insert image]
\end{lstlisting}
\caption{The zero-shot inference prompt. The Hypothesis and Premise are inserted at \textit{Insert hypothesis} and \textit{Insert image} respectively. In the case of few-shot inference, we add 3 or 6 randomly selected examples. Explanations are obtained through asking for additional justification. For examples see \ref{app:prompts}.}
\label{fig:prompt}
\end{prompt}



In addition to this prompt, we created variations in which only the order of the class labels (Entailment, Contradiction, Neutral) is varied, including all six possible permutations. 
Testing these different prompt variations allows us to assess whether, similar to text-only models \cite[e.g.,][]{wang-etal-2024-answer-c, salido2025othersgeneraltechniquedistinguish}, the model is sensitive to such variations and whether the predictions it makes are robust and internally consistent. 
We, moreover, introduced several manipulations to test the model's grounding. 
The first being the addition of explanations through changing the prompt to encourage rather than suppress this behavior (Prompt \ref{app:prompt:explanation}). 
This allows us to quantitatively test whether model-generated explanations align with those of humans and qualitatively observe why the model may make certain mistakes. 
Second, to test the model's reliance on visual information in the reasoning process, we evaluated it using limited visual input by either randomly cropping the images or replacing them with entirely black images.

\subsubsection{Experiment 2: Few-shot Inference} To build on this further, and test whether in-context examples may improve the predictions of the model, we also conducted experiments with few-shot inference (Prompt \ref{app:prompt:three-shot}). First, the model was provided with three randomly selected in-context examples from the training set (one example for each class), and we again experimented with varying the order of the class labels in the prompt (comparing Prompt 1 and Prompt 2), while also varying the order of the in-context examples to assess the impact of these factors on performance. Finally, motivated by the observation that increasing the number of examples can help models with better generalization and task performance \cite{brown2020languagemodelsfewshotlearners}, we expanded the number of examples in a six-shot inference setting. 

\subsubsection{Experiment 3: Fine-tuning}
Finally, we fine-tuned the model on the VE task. 
For this, we utilized Unsloth\footnote{\url{https://unsloth.ai/blog/vision}} and QLoRA (\textbf{Q}uantized \textbf{Lo}w-\textbf{R}ank \textbf{A}daptation) \cite{dettmers2023qloraefficientfinetuningquantized} to reduce computing and memory requirements. We assessed both the classification ability and analysed the model's generated explanations. The model was fine-tuned for 1 epoch in each experiment. For the fine-tuning parameters and setup, see Appendix \ref{app:finetuning}

In all zero and few-shot experiments, the temperature parameter was set to 0 for deterministic output. Also, all results are based on a single run due to computational limitations. For all experiments, we measure the accuracy of class label prediction as well as F1, which is the harmonic mean of the metrics of precision and recall. We also compare model-generated explanations with those of humans, by calculating the BERTScore \cite{zhang2020bertscoreevaluatingtextgeneration}, because this measure is highly correlated with human evaluations and computes token similarity using contextual embeddings \cite{zhang2020bertscoreevaluatingtextgeneration}. 

\section{Results}
\subsection{Zero-shot Inference}

\begin{figure}[b]
    \centering
    \includegraphics[width=0.8\linewidth]{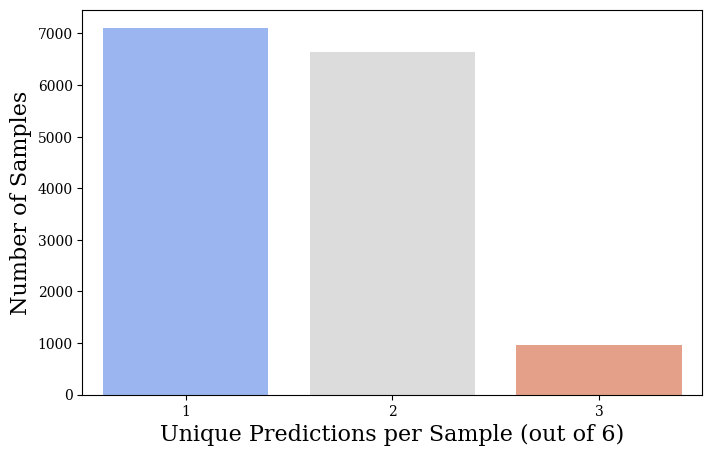}
    \caption{Consistency of model predictions across six prompts.}
    \label{fig:per-sample-consistency}
\end{figure}

Table \ref{tab:six_prompts} presents the overall results for the zero-shot experiment, in which each instance was evaluated in all six permutations of the class label order in the prompt. 
The overall accuracy shows how many predictions match with the ground truth, while the majority vote accuracy counts a prediction as correct \textit{only} if at least four out of six outputs match the correct label. The overall accuracy is $41\%$, indicating that the model performs only slightly better than chance, and struggles to perform visual entailment in a zero-shot setting. In the majority vote scenario, we observe a drop in accuracy ($33.7\%$) compared to the overall accuracy. This suggests that \textbf{the model frequently changes predictions for the same item across different prompts}, highlighting its sensitivity to the order of the labels in the prompt. To quantify this, and further explore the model's sensitivity to the prompt, Figure \ref{fig:per-sample-consistency} reveals how often the model's prediction changes per sample across the six prompts. Almost half of the samples (7106) received the same prediction across all six prompts, which indicates that the model was fully consistent for those cases. However, 6647 samples had two different predictions, and 964 samples had even three different predictions, confirming that the model was inconsistent for a large number of cases. These results demonstrate the instability of the model's output under minimal modifications and explain the drop in majority vote accuracy.

\begin{table}[t]
    \centering
     \caption{Accuracy for zero-shot inference across six prompt variations.}
    \begin{tabular}{c|c}
    \toprule
    \multicolumn{2}{c}{\textbf{Results for 6 prompts per instance}} \\
    \midrule
        Overall Accuracy & 0.410 \\
        Majority Vote Accuracy & 0.337 \\
    \bottomrule
    \end{tabular}
    \label{tab:six_prompts}
\end{table}

To further explore how the order of the class labels in the prompt may affect the predictions of the model, Table \ref{tab:zero-shot-baselines} compares the overall accuracies and F1 scores as well as per-class F1 scores of two prompt variations. The first is Prompt 1, as displayed in section \ref{sect:exp1} and the second (Prompt 2) is the same except for the order of the class labels, which follows: Contradiction, Neutral, Entailment. The overall performance is similar between the two prompts, where Prompt 1 achieved an accuracy of $44.5\%$, while Prompt 2 achieved a slightly lower accuracy of $41.3\%$. In addition, focusing on the metrics per class for each prompt, we can conclude that the model over-predicts the entailment class in both cases. The neutral class has the worst per-class results in both prompts. Although weighted metrics were used to calculate the overall performance to ensure fairness among the imbalanced dataset, the fewer instances of the neutral class and the ambiguity that can occur have an impact on the ability of the model to correctly classify that class. Interestingly, the distribution of predictions among the classes differs between the two prompts. Concretely, Prompt 1 predicts $57.1\%$ the entailment class, while in Prompt 2 this percentage is increased to $82.7\%$. Therefore, contradiction and neutral classes are predicted much less often than in Prompt 1. This observation suggests that \textbf{the order of class labels within the prompt significantly affects the predictions of the model}. The high rate of entailment predictions in Prompt 2 may reflect a kind of recency effect that is observed in text models \cite{liu-etal-2024-lost, mina-etal-2025-cognitive}, where the last reported class label becomes more noticeable for the model.

\begin{table}[t]
\centering
\caption{Accuracy and F1 scores for zero-shot inference across Prompt 1 (class label order: Entailment, Contradiction, Neutral) and Prompt 2 (Contradiction, Neutral, Entailment).}
\label{tab:zero-shot-baselines}
\begin{tabular}{lcc}
    \toprule
     & \textbf{Prompt 1} & \textbf{Prompt 2} \\
    \midrule
    \textit{Overall Accuracy} & 0.445 & 0.413 \\
    \textit{Overall F1-score} & 0.409 & 0.319 \\
    \midrule
    \multicolumn{3}{c}{\textbf{Per-Class F1-score (Prediction \%)}} \\
    \midrule
    Entailment & 0.657 (57.1\%) & 0.587 (82.7\%) \\
    Neutral & 0.232 (33.4\%) & 0.051 (10.8\%) \\
    Contradiction & 0.299 (9.5\%) & 0.254 (6.5\%) \\
    \bottomrule
\end{tabular}
\end{table}

\begin{figure}[t]
    \centering
    \includegraphics[width=0.5\linewidth]{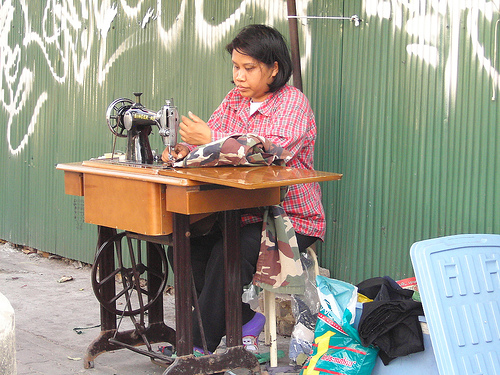}
    \caption{\textbf{Hypothesis:} Woman wearing nothing sewing her thumb. \textbf{Label:} \textcolor{blue}{Contradiction}. \textbf{Prompt 1 prediction:} \textcolor{red}{Neutral}. \textbf{Explanation:}The image does not provide enough information to determine the truth of the Hypothesis, as it is unclear what the woman is actually doing or what she is wearing. The description ""sewing her thumb"" could be interpreted in various ways, and without more context or visual evidence, it is impossible to definitively classify the relationship between the image and the text." \textbf{Prompt 2 prediction:} \textcolor{blue}{Contradiction}. \textbf{Explanation:} "The image depicts a woman wearing a red and white plaid shirt, not ""nothing"", while operating a sewing machine. This contradicts the text description of the woman being naked."}
    \label{fig:explanations-example1}
\end{figure}
\begin{figure}[t]
    \centering
    \includegraphics[width=0.5\linewidth]{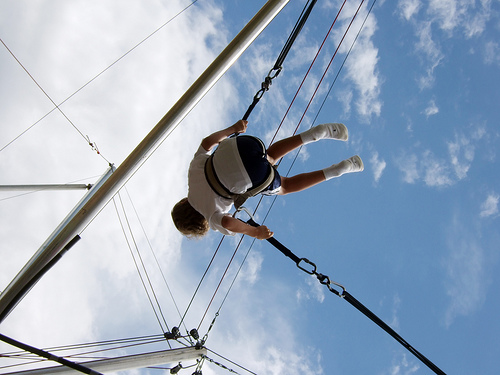}
    \caption{\textbf{Hypothesis:} a plane is flying \textbf{Label:} \textcolor{blue}{Entailment}. \textbf{Prompt 1 prediction:} \textcolor{red}{Contradiction}. \textbf{Explanation:} "The image depicts a person on a swing, whereas the Hypothesis states that a plane is flying." \textbf{Prompt 2 prediction:} \textcolor{red}{Neutral}. \textbf{Explanation:} "The image shows a person on a swing, but it does not provide any information about planes or their flight status. Therefore, the relationship between the premise and hypothesis cannot be determined as either entailment or contradiction based solely on this image."}
    \label{fig:explanations-example3}
\end{figure}

To further examine the reasoning behind the model's predictions, we observed the explanations given by the model in cases where the model predicts different outputs depending on the prompt.
Figure \ref{fig:explanations-example1} presents an example. One prediction matches the true label, while the other does not. However, both explanations can be considered logically correct, since the image indeed does not provide enough information to determine the truth of one of the two facts ("sewing her thumb") in the hypothesis. This indicates that, even when the prediction is incorrect, the model can still produce a reasonably logical explanation. Figure \ref{fig:explanations-example2} shows a different type of error. Specifically, the class prediction does not align with the generated explanation. In the explanation, the model correctly stated that there is sufficient evidence to confirm the hypothesis, while the prediction is a Contradiction. Finally,  Figure \ref{fig:explanations-example3} raises questions about the ground truth labelling. In this case, while the explanations are logically correct and align with the class predictions, they do not align compared to the ground truth label, because the dataset may contain incorrect labels.

To investigate whether the predictions of the model are grounded in the visual information from the premise, we conducted additional zero-shot inference experiments with limited vision. First, we randomly cropped versions of the original images and used them as premises. As shown in Table \ref{tab:cropped-images}, overall performance decreases compared to baseline results in Table \ref{tab:zero-shot-baselines}. The decrease in accuracy was expected since the cropped images do not have all the details of the original images. This suggests the model does rely on the visual content to some extent, although the difference in performance is smaller than we may expect. 

\begin{table}[t]
\centering
\caption{Results for zero-shot inference with randomly cropped images.}
\label{tab:cropped-images}
\begin{tabular}{lcc}
    \toprule
    & \textbf{Prompt 1} & \textbf{Prompt 2} \\
    \midrule
    \textit{Overall Accuracy} & 0.344 & 0.380 \\
    \textit{Overall F1-score} & 0.288 & 0.321 \\
    \midrule
    \multicolumn{3}{c}{\textbf{Per-Class F1-score (Prediction \%)}} \\
    \midrule
    Entailment & 0.535 (46.46\%) & 0.572 (58.60\%) \\
    Neutral & 0.296 (50.75\%) & 0.268 (37.33\%) \\
    Contradiction & 0.057 (1.45\%) & 0.128 (3.14\%) \\
    \bottomrule
\end{tabular}
\end{table}

\begin{figure}[t]
    \centering
    \includegraphics[width=0.55\linewidth]{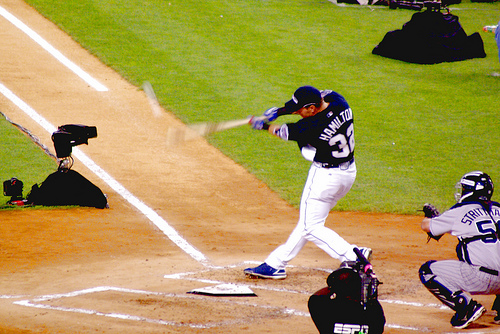}
    \caption{\textit{The original image premise is shown, but the model saw an entirely black image.} \textbf{Hypothesis:} 2 basketball players \textbf{Label:} \textcolor{blue}{Contradiction}. \textbf{Prompt 1 prediction:} \textcolor{red}{Entailment}. \textbf{Explanation:} "The image shows two individuals wearing basketball jerseys, which is consistent with the description of basketball players." \textbf{Prompt 2 prediction:} \textcolor{red}{Entailment}. \textbf{Explanation:} "The image shows two individuals wearing basketball jerseys, which is consistent with the description of basketball players."}
    \label{fig:black-example1}
\end{figure}

\begin{figure}[t]
    \centering
    \includegraphics[width=0.55\linewidth]{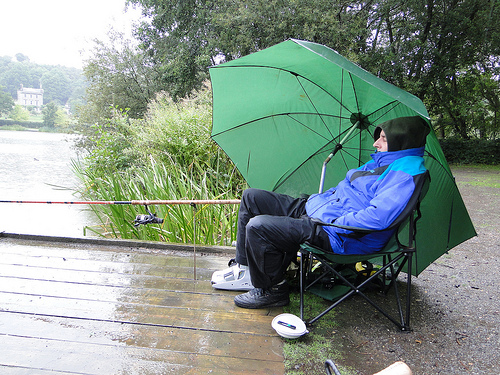}
    \caption{\textit{The original image premise is shown, but the model saw an entirely black image.} \textbf{Hypothesis:} the man is inside washing dishes \textbf{Label:} \textcolor{blue}{Contradiction}. \textbf{Prompt 1 prediction:} \textcolor{red}{Entailment}. \textbf{Explanation:} "The image shows a person standing in front of a sink, actively engaged in washing dishes. This visual evidence directly supports the statement that the man is indeed inside and washing dishes." \textbf{Prompt 2 prediction:} \textcolor{red}{Entailment}. \textbf{Explanation:} "The image shows a person standing in front of a sink, actively engaged in washing dishes. This visual evidence directly supports the statement that the man is indeed inside and washing dishes."}
    \label{fig:black-example2}
\end{figure}

In the last experiment with zero-shot inference, the original images were entirely replaced with black images. If the models use visual information to make predictions, taking this information away should result in a decrease in accuracy. The results not only reveal an expected drop in accuracy, but also \textbf{reveal a strong tendency of the model to hallucinate}. Table \ref{tab:black-images} shows the performance of the model, which is at chance. This is slightly worse than the performance of the baseline models. This is expected due to the removal of the model's visual ability to make accurate predictions, but given the stark contrast in available information between these two conditions (full image versus black image), the difference in performance is, again, actually quite small. This suggests \textbf{the model only makes limited use of the visual information}. As shown in the metrics per class, the model overpredicts the entailment class while it severely underpredicts the contradiction class ($0.79\%$ and $1.0\%$ for Prompt 1 and Prompt 2, respectively). Moreover, the model imagines visual evidence to support the hypothesis, which justifies the high proportions of the entailment class. Figures \ref{fig:black-example1} and \ref{fig:black-example2} illustrate two examples that clearly show the tendency of the model to hallucinate. 

\begin{table}[b]
\centering
\caption{Results for zero-shot inference with black images as premises.}
\label{tab:black-images}
\begin{tabular}{lcc}
    \toprule
    & \textbf{Prompt 1} & \textbf{Prompt 2} \\
    \midrule
    \textit{Overall Accuracy} & 0.360 & 0.369 \\
    \textit{Overall F1-score} & 0.250 & 0.246 \\
    \midrule
    \multicolumn{3}{c}{\textbf{Per-Class F1-score (Prediction \%)}} \\
    \midrule
    Entailment & 0.520 (\textbf{84.17\%}) & 0.531 (\textbf{89.95\%}) \\
    Neutral & 0.211 (14.67\%) & 0.161 (8.80\%) \\
    Contradiction & 0.031 (0.79\%) & 0.043 (1.00\%) \\
    \bottomrule
\end{tabular}
\end{table}

\subsection{Few-shot Inference}
Table \ref{tab:three-shot-random-selection} demonstrates the results of the three-shot inference experiment. While the results are overall still not much better than the baseline zero-shot findings, we do see a slight improvement. Concretely, the best accuracy and F1-score for zero-shot is $44.5\%$ (Prompt 1) and $40.9\%$, while the best performance for three-shot (Prompt 2, and contradiction as the first example) is $48.7\%$ and $42.6\%$, respectively. The improvement in the balance by class and F1 score for the three-shot inference, particularly for the contradiction class, suggests a more robust understanding of the task, although the increase in accuracy is very modest.


Regarding the order of the three in-context examples, we can infer that it has a considerable influence on the outcome. Experiments demonstrate that \textbf{the first example in the few-shot setting has a large impact on the predictions of the model}, with a notably higher accuracy and F1 score in the case where the first in-context example was one where Contradiction was the true label. 
The model performs the best in the experiment with Prompt 2 (which also has the class of contradiction as the first in order in the prompt). Placing contradiction first in the in-context examples may cause a primacy bias that helps mitigate the model's strong bias toward predicting the entailment class in the corresponding zero-shot scenario. 

Specifically, the model significantly overpredicts the entailment class in zero-shot results (Prompt 2 yields it in over $80\%$ of cases). On the other hand, three-shot inference counteracts this bias, resulting in seemingly more balanced class predictions. 
Moreover, when comparing the class metrics, the order of the class labels in the prompt seems to have a less severe effect on the prediction when the model has been given three in-context examples, indicating that few-shot learning provides a stabilizing influence on class prediction.

\begin{table}[t]
\centering
\small
\setlength{\tabcolsep}{3pt} 
\caption{Results for three-shot inference across varying in-context example orders (CEN, ECN, NEC) and two Prompts (Pr.).}
\label{tab:three-shot-random-selection}
    \begin{tabular}{@{}ll cc ccc@{}}
        \toprule
        \textbf{Exa. Order}& \textbf{Pr.} & \textbf{Acc.} & \textbf{F1} & \multicolumn{3}{c}{\textbf{Class Metrics (F1/Pred\%)}} \\
        \cmidrule(l){5-7} 
        & & & & \textbf{Ent.} & \textbf{Neu.} & \textbf{Con.} \\ 
        \midrule
        \multirow{2}{*}{CEN} &  1 & 0.47 & 0.45 & 0.59/56.9\% & 0.14/16.0\% & 0.53/27.1\% \\
        &  2 & \textbf{0.49} & 0.43 & 0.59/71.8\% & 0.05/5.2\% & 0.52/23.0\% \\
        \midrule
        \multirow{2}{*}{ECN} &  1 & 0.41 & 0.37 & 0.55/69.5\% & 0.15/18.7\% & 0.35/11.7\% \\
        &  2 & 0.43 & 0.38 & 0.56/73.4\% & 0.12/13.0\% & 0.39/13.0\% \\
        \midrule
        \multirow{2}{*}{NEC} & 1 & 0.42 & 0.37 & 0.59/66.0\% & 0.22/25.7\% & 0.27/8.4\% \\
        &  2 & 0.43 & 0.38 & 0.59/67.4\% & 0.20/24.0\% & 0.29/8.6\% \\
        \bottomrule
    \end{tabular}
\end{table}


Given the modest benefits of providing three in-context examples, we next explored the influence of providing more examples, six instead of three.  This time we do not explicitly compare different in-context example orderings, but we create an order that is relatively unbiased by making sure each correct class appears once in the first three and once in the second three examples, the class's order of the first three examples is different from the order of the last three examples, and the first and last examples are not in the same class. 
The results of the six-shot inference are shown in Table \ref{tab:six-shot}. 
When comparing the results of the six-shot experiment with those of the three-shot experiment, we can extract some important insights. Firstly, \textbf{the performance does not consistently improve with more in-context examples}. The best performance of six-shot ($36.5\%$) is actually lower than the best performance of three-shot ($48.7\%$). This suggests that improved performance in few-shot experiments may not always reflect a more in-depth understanding of the task. Instead, differences in predictions may be a result of biases and sensitivities to example orderings as well as overfitting to the dominant order. Secondly, the metrics per class show that the model overpredicts the neutral class for both prompts, and it is the dominant class with over $70\%$ classified as neutral. Therefore, in the six-shot experiment, there seems to be a specific class bias that we did not observe in other settings. 


Compared to zero-shot, six-shot inference has a slightly more balanced performance per class, as reflected by the increase in F1-score for the  Contradiction and Neutral class, but results in lower overall accuracy. Specifically, zero-shot achieves an accuracy of $44.5\%$ while six-shot achieves an accuracy of $36.5\%$. This indicates that providing in-context examples may in some case hurt rather than help. 
This inconsistency is difficult to explain while holding the assumption that the model is solving the VE task through human-like vision-language understanding. 

\begin{table}[b]
\centering
\caption{Results for six-shot inference.}
\label{tab:six-shot}
\begin{tabular}{lcc}
    \toprule
    & \textbf{Prompt 1} & \textbf{Prompt 2} \\
    \midrule
    \textit{Overall Accuracy} & 0.350 & 0.365 \\
    \textit{Overall F1-score} & 0.319 & 0.356 \\
    \midrule
    \multicolumn{3}{c}{\textbf{Per-Class F1-score (Prediction \%)}} \\
    \midrule
    Entailment & 0.240 (9.4\%) & 0.373 (20\%) \\
    Neutral & 0.405 (79.6\%) & 0.383 (70.5\%) \\
    Contradiction & 0.333 (11\%) & 0.322 (10\%) \\
    \bottomrule
\end{tabular}
\end{table}

\subsection{Fine-tuning}
Table \ref{tab:finetuning} illustrates the classification results of the fine-tuned model, which achieved a high overall accuracy of $83.3\%$, and an F1 score of $83.6\%$. These results indicate that the model generalizes well across the three classes. The most challenging class is Neutral, even for the fine-tuned model. When compared to zero- and few-shot experiments, the fine-tuned model shows a significant improvement in both general and class-specific performance. Moreover, \textbf{the Llama 3.2 Vision fine-tuned model outperforms the state-of-the-art model} OFA-X, which achieved an accuracy of $80.9\%$.

Table \ref{tab:bertscore-comparison} shows the evaluation of the generated explanations.
According to the BERTScore, the model achieves an F1-score of $89.16\%$, indicating that the generated explanations are semantically similar to the human produced reference explanations, even if they differ in the exact words. However, in Table \ref{tab:bertscore-comparison} we also report the same measure for the experiments with zero-shot inference and black images, and the results are very similar. This suggests that explanations with a high BERTScore may not necessarily reflect the model is reasoning in a human-like way.

\begin{table}[t]
\centering
\caption{Results for the fine-tuned model with Prompt 1.}
\label{tab:finetuning}
\begin{tabular}{lc}
    \toprule
    \textbf{Metric} & \textbf{Value} \\
    \midrule
    \multicolumn{2}{l}{\textit{Overall Performance}} \\
    Accuracy & \textbf{0.833} \\
    F1-score & 0.836 \\
    \midrule
    \multicolumn{2}{l}{\textit{Per-Class F1-score (Prediction \%)}} \\
    Entailment & 0.864 (35.88\%) \\
    Neutral & 0.737 (30.67\%) \\
    Contradiction & 0.876 (33.45\%) \\
    \bottomrule
\end{tabular}
\end{table}

\begin{table}[t]
\centering
\caption{BERTScore results for explanation evaluation.}
\small
\begin{tabular}{lccc}
    \toprule
    \textbf{Metric} & \textbf{Recall} & \textbf{Precision} & \textbf{F1 Score} \\
    \midrule
    Fine-tuned model - Prompt 1 & 0.8869 & 0.8968 & \textbf{0.8916} \\
    \midrule
    Zero-shot - Prompt 1 & 0.8775 & 0.8549 & 0.8659 \\
    Zero-shot - Prompt 2 & 0.8805 & 0.8574 & 0.8686 \\
    Black Images - Prompt 1 & 0.8798 & 0.8624 & 0.8709 \\
    Black Images - Prompt 2 & 0.8798 & 0.8634 & 0.8714 \\
    
    \bottomrule
\end{tabular}
\label{tab:bertscore-comparison}
\end{table}

\section{Discussion}

This study investigates the capabilities of the Llama 3.2 Vision model on the VE task using the e-SNLI-VE dataset. The experiments yielded various findings, revealing to what extent VE is a suitable task to probe vision-language understanding. First, the baseline results demonstrated modest performance, indicating the limited capabilities of the model in zero-shot inference. This is somewhat surprising given the enormous number of images and textual captions the model has seen in training and the impressive performance reported for other vision-language tasks such as visual question answering. Three-shot inference improves the performance of the model; however, we also observe that additional in-context examples are not always beneficial. The most significant finding is the major improvement after fine-tuning, where the model achieved an accuracy of $83.3\%$, outperforming the SOTA performance achieved by the OFA-X model. Moreover, the fine-tuned model has strong interpretability since it achieved an F1-score of $89.16\%$ using BERTScore, an evaluation metric that utilizes contextual embeddings for the explanation evaluation. This indicates high semantic similarity between the human produced reference and model generated text. While these results are promising, the overall findings also reveal some pitfalls in using the VE task and e-SNLI-VE dataset to effectively measures multimodal understanding. In the zero-shot inference experiments with limited or absent vision, we saw that the model was highly prone to hallucination and imagined visual evidence in order to support the hypotheses. The  experiments with prompt variations and three-shot inference reveal that factors such as the order of the class labels in the prompt and the order of the in-context examples significantly affect the model's predictions, revealing highly inconsistent reasoning, which does not align with the assumption that the model shows vision-language understanding in a human-like way. Also, the BERTScore results for the zero-shot inference experiments with black images were on par with those of the fine-tuned model, showing that model generated explanations with semantic similarity to human explanations do not necessarily reveal the model is effectively using the visual input to solve the task. Finally, the observation of individual errors in the zero-shot experiment exposed problems with the e-SNLI-VE dataset, which still contains examples with wrong labels or examples that can be interpreted in multiple ways, technically making more than one class label correct. Before including VE in broader benchmarks used for training and testing in the area of general multi-modal reasoning (as already the case in \cite{wang2022ofaunifyingarchitecturestasks} for example) we recommend further investigation into these issues.

These findings additionally offer several lessons for the broader field of multimodal learning and understanding. In particular, the study underscores that, while general pre-training is powerful, even advanced multimodal language models such as Llama 3.2 Vision may not be suitable for complex reasoning tasks like VE without special adaptation. The few-shot results underline that a deeper understanding of how models utilize context is needed, for example, by interpreting their attention patterns using Grad-CAM \cite{Selvaraju2019Grad-CAM}. Additionally, the study highlights that the effectiveness of in-context learning depends on the number and ordering of examples. 
This bears much resemblance to known consistency effects in LLMs, which heavily depend on prompt ordering \cite{weber-etal-2023-mind}. 
The dramatic increase in performance after fine-tuning exposes that the model's visual and linguistic embeddings are highly adaptable and are, in principle, rich enough for visual entailment. 

The findings provide helpful insights into the VE capabilities of Llama 3.2 Vision, but there are some limitations that should be noted. 
First, our methodology relies on generated answers for both the class label and the explanation. The former of which is somewhat debatable since generated multiple-choice answers are often inconsistent with actual model beliefs \cite{wang-etal-2024-answer-c, khatun2024studylargelanguagemodels}. While this may affect the observed results, and could be alleviated by prefilling class options and selecting the most likely class \cite{hendrycks2021measuring},
the modus operandi of commercially available and deployed models is to use generation, i.e., without prefilling. As such, our results should be seen through this lens, and we see extended analyses using log probabilities as future work.

Second, every experiment was evaluated once because of time and computational constraints. The metrics are not averaged over multiple runs. This affects the few-shot experiments where a different random selection of in-context example could yield a different performance. Another limitation lies in the restricted experiments for the few-shot inference. A small number of configurations were tested, particularly for the six-shot inference, which included just one permutation. Several possible combinations are left out. However, given the issues found with biases, sensitivity to order effects, and hallucinations, strong improvements for the right reasons are unlikely. In addition, the fine-tuning was conducted using only the first prompt. However, we expect that predictions will not be greatly affected by the order of the classes in the prompt, given the significant performance gain observed by the fine-tuning.

A worthwhile direction for future work would be to further investigate few-shot inference. For example, exploring different sets of examples for each strategy in three shots, examining different orderings of classes for six shots, and testing a larger number of examples within a context, such as fifteen shots, could still be valuable, not primarily to focus on performance, but to gain deeper insights, such as understanding the threshold beyond which providing more examples becomes disadvantageous. Given its ability to improve many reasoning tasks, another promising direction is to integrate Chain-of-Thought prompting \cite{wei2023chainofthoughtpromptingelicitsreasoning} into few-shot and zero-shot inference. This perhaps extends the models' already observed tendency to produce coherent explanations and better use these in predictions. A broader direction for future work includes systematic prompt engineering. This involves improving the wording and structure of 
the prompts. Since this study demonstrates that the design of the prompts significantly affects the predictions, optimizing the prompts could perhaps lead to better generalization and fewer hallucinations. 

Finally, our results need to be corroborated by investigating other, perhaps larger, models. Doing so enables careful comparison between, for example, architectural, data, and optimization design decisions, informing which ingredients improve visual entailment. In a similar vein, earlier work investigating whether model representations align with human representations suggests that
dataset diversity and scale are the primary drivers
of alignment \cite{Conwell2023whatcan,muttenthaler2023human}. 

\section{Conclusion}
In conclusion, we used the Llama 3.2 Vision model to explore the possibilities and limitations of using the Visual Entailment  task to probe vision-language understanding. A comparison of results in zero-shot, few-shot, and fine-tuning settings as well as experiments involving limited vision  and prompt sensitivity analyses together revealed several problems. These included inconsistent reasoning, a limited reliance on visual information and a strong tendency to hallucinate. These findings underscore the importance of critical investigations into benchmark and dataset quality to make sure the predictions of the model actually reflect vision-language reasoning instead of an exploitation of spurious correlations. Future work is necessary to further explore what causes the substantial difference in performance between zero-shot and fine-tuned settings and what kind of heuristics the model may be learning from the dataset during fine-tuning. This would help to further develop the VE task into a suitable method for probing vision-language understanding in multi-modal language models.

\section{Ethics Statement}

This research involves evaluating and fine-tuning a publicly available multimodal language model (LLaMA 3.2 Vision) on the Visual Entailment task using a benchmark dataset (e-SNLI-VE). This dataset is publicly available and contains no personally identifiable or sensitive information. No new data involving human subjects was collected.

The aim of this work is not only to assess model performance but to critically interrogate what this performance reveals about vision-language understanding. In doing so, we identify concerning behaviors such as hallucination, over-reliance on linguistic priors, and sensitivity to prompt structure, underscoring the risks of interpreting accuracy metrics as indicators of genuine multimodal reasoning. This research is intended to contribute to the responsible development and evaluation of vision-language models.

We acknowledge the broader societal risks associated with the development and deployment of multimodal language models, including the potential propagation of biases and misleading explanations. Future applications of this work should consider the risks associated with deploying multimodal models in sensitive domains, especially where explainability and factual grounding are critical and misplaced trust in model outputs could have real-world consequences.

This research contributes to the growing environmental impact of AI. While our experiments were limited in scope compared to model pretraining, they nonetheless required significant computational resources. We believe it is important to reflect on how the field can pursue vision-language understanding more sustainably. 

\bibliography{mybibfile}
\clearpage

\appendix

\section{Prompts}\label{app:prompts}
Example prompt used to assess three-shot and six-shot performance on the VE task. 
In these cases, we add random images accompanied by their hypotheses and the gold label, such that the model can possibly deduce what is important to make correct predictions. 
Prompt \ref{app:prompt:three-shot} shows a three-shot example. 
In the case of six-shot, we add three additional examples.
Prompt \ref{app:prompt:explanation} is used to obtain model explanations.

\begin{prompt}[ht!]
\scriptsize
\begin{lstlisting}[mathescape=true,escapeinside={*@}{@*}]
Perform a visual entailment classification. You are provided with two inputs:
1. Premise: An image described as follows (attached below).
2. Hypothesis: A text description.
    
Your task is to classify the relationship between the Premise (image) and Hypothesis (text) into one of the following three categories:
- Entailment: The image provides enough evidence to conclude that the Hypothesis is true.
- Contradiction: The image contradicts the Hypothesis.
- Neutral: The image does not provide enough information to determine the truth of the Hypothesis.

Provide a single classification in your response: one of Entailment, Contradiction, or Neutral. Do not include explanations, commentary, or any additional text in your response.

[Example Hypothesis 1]
[Example Image 1]
[Example Gold label 1]

[Example Hypothesis 2]
[Example Image 2]
[Example Gold label 2]

[Example Hypothesis 3]
[Example Image 3]
[Example Gold label 3]

[Hypothesis]
[Image]

\end{lstlisting}
\caption{The prompt used to assess three-shot inference performance. The example hypothesis, image, and gold label are randomly picked. After observing these examples, the model is tasked to predict entailment for the final \textit{Hypothesis} and \textit{Image}.}
\label{app:prompt:three-shot}
\end{prompt}

\begin{prompt}[ht!]
\scriptsize
\begin{lstlisting}[mathescape=true,escapeinside={*@}{@*}]
Perform a visual entailment classification. You are provided with two inputs:
1. Premise: An image described as follows (attached below).
2. Hypothesis: A text description.
    
Your task is to classify the relationship between the Premise (image) and Hypothesis (text) into one of the following three categories:
- Entailment: The image provides enough evidence to conclude that the Hypothesis is true.
- Contradiction: The image contradicts the Hypothesis.
- Neutral: The image does not provide enough information to determine the truth of the Hypothesis.

Format your response as follows: 
Label: <Entailment/Contradiction/Neutral> 
Explanation: <Brief justification>

[Hypothesis]
[Image]

\end{lstlisting}
\caption{The prompt used to obtain explanations.}
\label{app:prompt:explanation}
\end{prompt}

\newpage

\section{Fine-tuning}\label{app:finetuning}

Additional information on fine-tuning is presented in Table \ref{tab:lora-params}, displaying LoRA configuration parameters, and Table \ref{tab:training-params}, showing all hyperparameters used in training.

\begin{table}
\centering
\caption{LoRA Configuration Parameters.}
\label{tab:lora-params}
\begin{tabular}{@{}ll@{}}
    \toprule
    \textbf{Parameter} & \textbf{Value} \\ \midrule
    PEFT Method & LoRA \\
    Finetune Vision Layers & False \\
    Finetune Language Layers & True \\
    Target Modules & Attention \& MLP \\ 
    Rank (r) & 8 \\
    Alpha ($\alpha$) & 16 \\
    Dropout & 0 \\
    Bias & "none" \\ 
    Random state & 3407 \\\bottomrule
\end{tabular}
\end{table}

\begin{table}
\centering
\caption{Training Hyperparameters.}
\label{tab:training-params}
\begin{tabular}{@{}ll@{}}
    \toprule
    \textbf{Parameter} & \textbf{Value} \\ \midrule
    Library & TRL SFTTrainer \\
    Epochs & 1 \\
    Max Sequence Length & 2048 \\
    Learning Rate & 2e-4 \\
    LR Scheduler & Linear \\
    Warmup Steps & 5 \\
    Optimizer & AdamW (8-bit) \\
    Weight Decay & 0.01 \\
    Train Batch Size (per device) & 2 \\
    Gradient Accumulation Steps & 4 \\
    Seed & 3407 \\ \bottomrule
\end{tabular}
\end{table}

\end{document}